\documentclass{article}
\usepackage{spconf,amsmath,graphicx,hyperref}
\usepackage[capitalize,noabbrev]{cleveref}

\usepackage{amsmath}
\usepackage{amssymb}
\usepackage{mathtools}
\usepackage{amsthm}
\usepackage{float}
\usepackage{placeins}

\usepackage[textsize=tiny]{todonotes}

\usepackage{multirow}
\usepackage{colortbl} 
\usepackage{xspace}
\usepackage{tcolorbox}
\usepackage{listings}
\usepackage{subcaption}
 \usepackage{booktabs}
 
\definecolor{shadecolor}{RGB}{220,220,220} 
\definecolor{customgreen}{RGB}{230, 248, 224} 
\definecolor{customblue}{RGB}{212, 230, 241} 
\newcommand{\Ours}[0]{\textit{TEACH}\xspace}

\title{TEACH: Text Encoding as Curriculum Hints for Scene Text Recognition}
\name{Xiahan Yang, Hui Zheng}
\address{ China Mobile Research Institute, Beijing, China\\Email: \{yangxiahan, zhenghui\}@chinamobile.com}
\begin{document}
%
\maketitle
\begin{abstract}
Scene Text Recognition (STR) remains a challenging task due to complex visual appearances and limited semantic priors. We propose \Ours, a novel training paradigm that injects ground-truth text into the model as auxiliary input and progressively reduces its influence during training. By encoding target labels into the embedding space and applying loss-aware masking, \Ours simulates a curriculum learning process that guides the model from label-dependent learning to fully visual recognition. Unlike language model-based approaches, \Ours requires no external pretraining and introduces no inference overhead. It is model-agnostic and can be seamlessly integrated into existing encoder-decoder frameworks. Extensive experiments across multiple public benchmarks show that models trained with \Ours achieve consistently improved accuracy, especially under challenging conditions, validating its robustness and general applicability.
\end{abstract}

\begin{keywords}
Scene Text Recognition, Curriculum Learning, Multimodal Learning, Vision-Language Models
\end{keywords}

\section{Introduction}
\label{sec:intro}

Scene Text Recognition (STR), a key task in Optical Character Recognition (OCR), aims to transcribe text from natural scene images. Compared to document OCR, STR involves a broader range of challenges, including perspective distortions, background clutter, uneven lighting, occlusions, and variations in fonts, styles, and languages. These conditions make text often ambiguous at the visual level, especially when the semantic context is weak or unavailable. While recent models have achieved strong results on curated benchmarks, their robustness in unconstrained real-world settings remains limited. Addressing these challenges requires models that can effectively extract reliable visual features without over-reliance on language priors.


\begin{figure}[tb]
    \centering
    \includegraphics[width=0.7\linewidth]{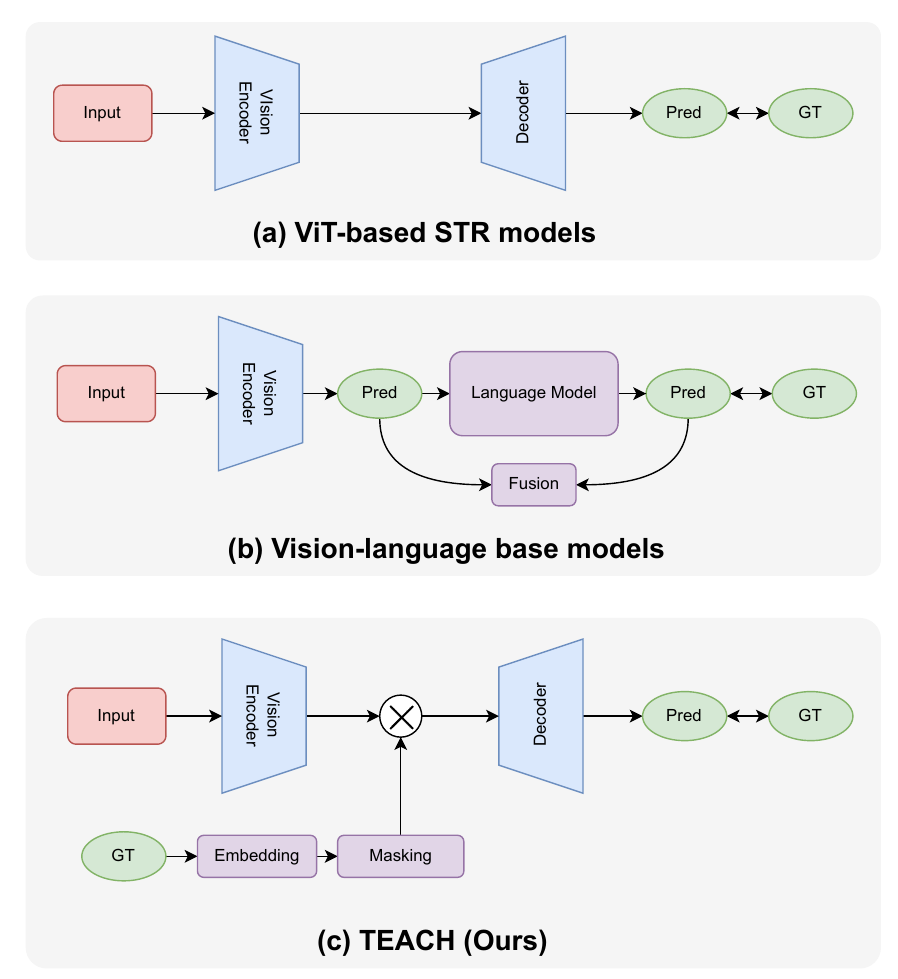}
\caption{
Comparison of different scene text recognition (STR) model paradigms. 
(a) ViT-based STR models directly predict character sequences from visual features using an encoder-decoder structure. 
(b) Vision-language models incorporate a separate language model and perform prediction through multimodal fusion, often requiring large-scale pretraining. 
(c) \Ours introduces ground-truth label embeddings into the training phase with progressive masking, guiding the decoder from label-conditioned prediction to vision-only recognition, without modifying inference-time structure.
}
    \label{fig:intro}
\end{figure}

As STR tasks grow in complexity, traditional models like the Convolutional Recurrent Neural Network (CRNN)~\cite{shi2016end_CRNN} struggle with challenging samples, especially those involving irregular layouts or long sequences. Recent advances leverage attention mechanisms~\cite{vaswani2017attention} and Vision Transformers (ViTs)~\cite{dosovitskiy2020image_vit}, which bring powerful global modeling capabilities. Transformer-based STR models such as ViTSTR~\cite{atienza2021vision_ViTSTR} (Figure~\ref{fig:intro} (a)) and PARSeq~\cite{bautista2022PARSeq} (See Figure~\ref{fig:intro} (b)) improve recognition accuracy by capturing long-range dependencies and enabling parallel decoding. These models significantly outperform RNN-based architectures on standard benchmarks. However, in the absence of strong language cues, their performance remains limited by the difficulty of learning robust visual features under noisy or ambiguous input conditions.

To alleviate this, multimodal approaches introduce external language information to provide additional semantic guidance~\cite{li2023trocr, fang2021read_ABInet, fujitake2024dtrocr, zhao2023clip4str}. Notable examples include ABINet~\cite{fang2021read_ABInet}, which fuses visual and textual streams in a bidirectional iterative process, and TrOCR~\cite{li2023trocr}, which employs pretrained language models as decoders. These methods achieve strong results by injecting semantic priors during decoding. However, they often rely on large-scale pretrained language models, increasing model size, training complexity, and inference latency. Moreover, their effectiveness can be fragile when the input text is out-of-distribution or lacks semantic coherence, as these models tend to overfit to seen language patterns rather than relying on visual cues. This raises a critical question: can we guide the model more effectively during training—without relying on external language models—so that it learns to extract reliable visual representations even under weak or noisy contexts?

Our method is motivated by how humans learn to read. In early stages, character recognition typically precedes full word understanding, and visual patterns are prioritized over semantic reasoning. For instance, when reading the word “TYPO,” a reader relying solely on visual familiarity can correctly recognize it, while one driven by linguistic priors may incorrectly infer it as “TYPE.” This suggests that language priors—while helpful—can also introduce bias, especially when texts deviate from semantic norms. In such cases, over-reliance on pretrained language models may degrade recognition accuracy. Instead, effective reading relies on gradually developing visual discrimination capabilities through guided exposure and repetition. This process is consistent with the principles of curriculum learning~\cite{bengio2009curriculum} and self-paced learning~\cite{kumar2010self}, where simpler or more informative signals are introduced early and gradually removed as the learner gains competence. Inspired by this, we design \Ours to inject explicit textual supervision in the early training phase and progressively reduce its influence, allowing the model to learn robust visual representations without depending on external language models.

Building on this intuition, we propose \Ours, a training strategy that simulates the staged nature of human reading acquisition. Instead of relying on external language models, \Ours introduces ground-truth text directly into the model input during training, providing explicit supervision to guide the learning of visual representations (See Figure~\ref{fig:intro} (c)). As training progresses, this guidance is gradually reduced through a loss-aware masking mechanism, enabling the model to rely increasingly on visual features. This process encourages the model to build up character-level recognition capabilities by aligning text embeddings with visual patterns—mirroring how humans gradually internalize character forms through repeated exposure and comparison. Unlike prior methods that depend on pretrained language priors, \Ours achieves this effect entirely within a vision-centric framework, without increasing inference cost or model size. Our experimental results demonstrate that this approach enhances model robustness and generalization, especially in challenging STR scenarios.

\noindent\textbf{Our main contributions are summarized as follows:}
\begin{itemize}
    \item We propose \Ours, a novel training paradigm for scene text recognition that injects ground-truth text into the input sequence and progressively masks it based on the model’s learning status.
    \item \Ours simulates a curriculum-style learning process inspired by human reading acquisition, enabling the model to transition from label-guided to vision-only recognition without relying on external language models.
    \item \Ours is model-agnostic, incurs no inference overhead, and consistently improves STR performance across multiple benchmarks, especially under challenging real-world conditions.
\end{itemize}

\section{Related Work}
\label{sec:relatework}

Scene Text Recognition (STR) methods can be broadly grouped into three major paradigms: convolutional-recurrent architectures, transformer-based models, and vision-language multimodal approaches. 

\begin{figure*}[t]
  \centering
  \includegraphics[scale=0.7]{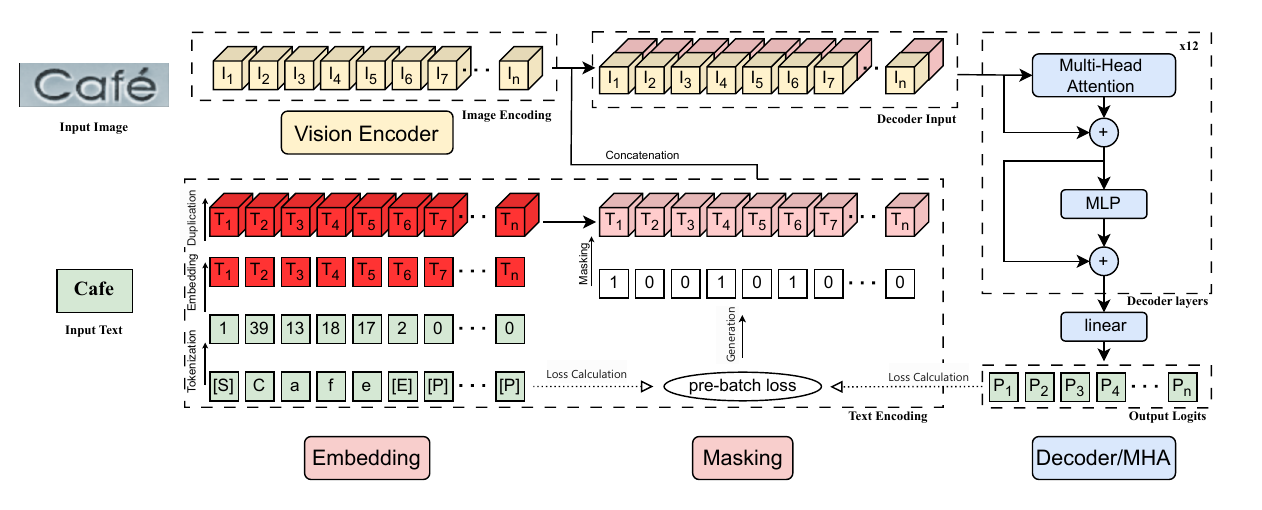}
\caption{
Training phase of \Ours. Ground-truth text is first embedded and aligned with the visual sequence output from the vision encoder. The two modalities are concatenated and passed to the decoder for prediction. A loss-aware masking module dynamically controls the amount of label information injected, gradually reducing textual guidance as the model converges. During inference, the model reverts to its original visual-only architecture without requiring any label input.
}

  \label{Fig:framework}
\end{figure*}

\textbf{CRNN and Recurrent Models.}
Convolutional Recurrent Neural Networks (CRNN)~\cite{shi2016end_CRNN} are early yet influential architectures in STR. They use CNNs to extract visual features, followed by RNNs (e.g., LSTMs, GRUs) for sequence modeling and CTC-based decoding. Variants such as R2AM~\cite{lee2016recursive_R2AM}, GCRNN~\cite{elbasani2021gcrnn}, and MA-CRNN~\cite{tong2020ma_MACRNN} enhance visual modeling with attention mechanisms and multi-scale designs.
Despite their success, these models suffer from several limitations: (1) RNNs are inherently sequential, which leads to slow inference and difficulty in parallel training; (2) long-range dependencies are hard to capture, often leading to gradient vanishing/explosion; (3) context modeling is sensitive to reading direction, and bidirectional RNNs only partially mitigate this. Moreover, performance degrades significantly in complex visual scenes due to limited global context modeling.

\textbf{Transformer-based STR Models.}
Transformers bring global receptive fields and parallel decoding to STR. ViTSTR~\cite{atienza2021vision_ViTSTR} employs a pure Vision Transformer encoder and a simple linear decoder, enabling efficient training and competitive accuracy. PARSeq~\cite{bautista2022PARSeq} further introduces Permuted Sequence Modeling (PSM), which creates random token orders to better learn non-monotonic character dependencies and reduce reliance on left-to-right prediction.
These transformer-based models significantly outperform RNN-based methods in both accuracy and speed. However, their reliance on vision-only signals makes them sensitive to noisy or low-quality images, especially in cases of curved text, occlusion, or cluttered backgrounds. Moreover, without strong semantic priors, transformers can mispredict under ambiguous conditions. While their architectural simplicity is an advantage, it also limits their ability to recover from uncertain or missing visual cues.

\textbf{Vision-Language Multimodal Models.}
To overcome the limitations of purely visual models, recent work has integrated textual semantics through multimodal fusion. ABINet~\cite{fang2021read_ABInet} separates visual and linguistic processing using a bi-directional completion network (BCN), and refines predictions iteratively through fusion. TrOCR~\cite{li2023trocr} and Clip4STR~\cite{zhao2023clip4str} leverage large pretrained language models (e.g., RoBERTa, CLIP) to guide decoding, while DTrOCR~\cite{fujitake2024dtrocr} simplifies the visual encoder and uses GPT-2 as a standalone decoder.
These models achieve state-of-the-art accuracy across benchmarks by injecting rich language priors. However, they come with several trade-offs: (1) pretrained LLMs significantly increase model size and training cost; (2) inference latency rises due to autoregressive decoding; (3) reliance on language priors can cause overfitting to training distributions and poor generalization on rare or out-of-vocabulary words. Models may incorrectly "correct" visually accurate but semantically uncommon words (e.g., reading “TYPO” as “TYPE”). Additionally, as shown in~\cite{bautista2022PARSeq}, frozen LLMs are unable to exploit ground-truth labels effectively without full fine-tuning—an expensive and often inefficient process.

\textbf{Positioning of Our Work.}
In contrast to models that rely on language priors at inference, our method \Ours injects label supervision only during training via a masked curriculum strategy. This enables the model to benefit from explicit textual guidance early on while progressively learning to rely solely on visual features. Unlike LLM-based models, \Ours adds no inference-time complexity, is model-agnostic, and improves robustness across challenging STR cases such as low resolution, irregular layouts, or missing semantic cues. Our approach shifts the role of text from post-hoc correction to proactive feature learning.

\section{Methods} \label{sec:method}

We propose \Ours, a plug-and-play training strategy for scene text recognition (STR) that injects ground-truth label embeddings into the model input and progressively removes them during training. This simulates a curriculum-style learning process where the model first relies on explicit textual supervision, then gradually learns to infer based purely on visual cues. \Ours is lightweight, model-agnostic, and requires no architectural changes or inference-time overhead. This section describes the overall framework and details the progressive learning process.

\subsection{Overview of \Ours}

As illustrated in \cref{Fig:framework}, \Ours operates within a standard encoder-decoder framework. During training, we encode the target text and concatenate it with visual features extracted from the image. To avoid overfitting to label tokens, a loss-aware masking strategy gradually suppresses the injected textual information. At inference time, no label is needed—the model behaves identically to its original design.

\noindent\textbf{Visual Encoder.}
\Ours does not alter the original image encoder, which can be any vision backbone (e.g., ViT, ResNet, or CNN). The encoder extracts a sequence of visual tokens $\textbf{F} \in \mathbb{R}^{L_i \times E_s}$ from the input image $img \in \mathbb{R}^{H \times W \times C}$, where $L_i$ is the number of image patches and $E_s$ is the embedding dimension. These tokens are passed to the decoder along with the encoded label embeddings.

\noindent\textbf{Text Encoding.}
Given a label sequence $X = \{x_1, x_2, \ldots, x_n\}$, we encode it into $\textbf{X}_{\text{input}} \in \mathbb{R}^{L_t \times E_s}$ using a dictionary lookup or a trainable embedding layer. At each training step, we compute the injected label signal $\tilde{\textbf{X}}$ (possibly masked) and concatenate it with visual tokens:
\begin{equation}
\textbf{y} = \text{Decoder} \left( [\textbf{F}; \tilde{\textbf{X}}] \right),
\end{equation}
where $[\cdot;\cdot]$ denotes concatenation along the sequence dimension.

\subsection{Progressive Training Process}

\Ours divides training into three stages—initial, intermediate, and final—based on the degree of dependence on label supervision.

\noindent\textbf{Initial Stage: Full Label Guidance.}
At the beginning of training, the image encoder is randomly initialized and may produce noisy or uninformative features. To provide strong early supervision, we inject the full label embedding into the decoder without masking:
\begin{equation}
\tilde{\textbf{X}} = \textbf{X}_{\text{input}}.
\end{equation}
This setup allows the decoder to focus on reconstructing the original label sequence directly from its embedding. Conceptually, the decoder is learning an approximate inverse of the text encoding function, denoted as $\mathcal{T}^{-1}(\cdot)$:
\begin{equation}
\textbf{y} \approx \mathcal{T}^{-1}(\textbf{X}_{\text{input}}).
\end{equation}
This stage effectively bootstraps the training process by aligning the embedding space with the output space, enabling rapid convergence before meaningful visual features are available.

\begin{figure*}[t]
  \centering
  \includegraphics[scale=0.7]{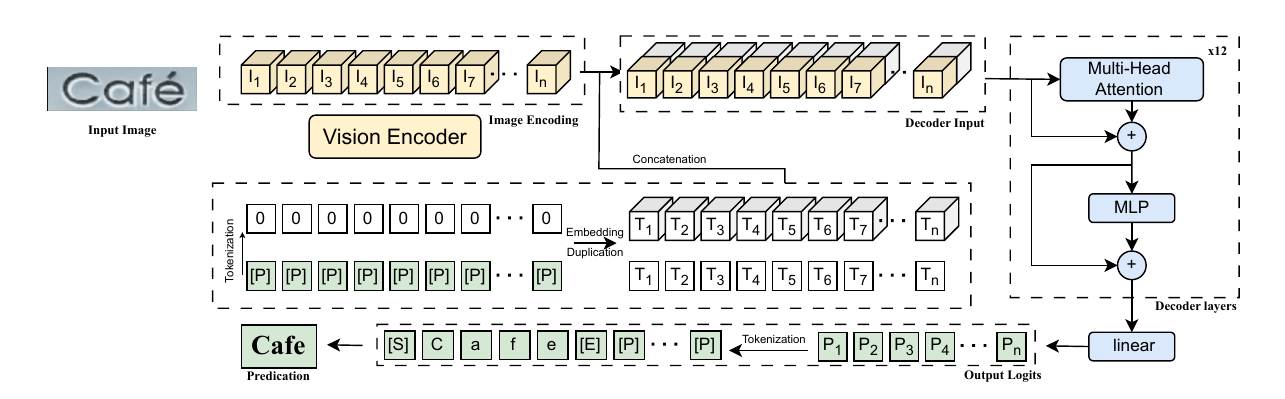}
\caption{
Inference phase of \Ours (TEACH). During testing, the model receives only the input image, and no textual supervision is provided. The vision encoder extracts image features, which are directly fed into the decoder. The decoder predicts the character sequence in an autoregressive manner, using special tokens for start ([S]) and padding ([P]). TEACH introduces no additional computation or input during inference, maintaining the efficiency and simplicity of the original STR model.
}
  \label{Fig:teach-test}
\end{figure*}

\noindent\textbf{Intermediate Stage: Loss-Aware Masking.}
As training proceeds and the model begins extracting meaningful visual features, we gradually reduce the influence of text input using a binary mask $\mathbb{M} \in \{0,1\}^{L_t}$:
\begin{equation}
\tilde{\textbf{X}} = \mathbb{M} \odot \textbf{X}_{\text{input}} + (1 - \mathbb{M}) \odot pad_{id},
\end{equation}
where $\odot$ denotes element-wise multiplication and $pad_{id}$ is a fixed or learnable padding vector. The masking rate $r$ is dynamically computed based on the previous batch’s loss:
\begin{equation} \label{eq:mask_ratio}
r = \max(0, \min(1, \alpha \cdot (\text{Loss} - \beta))),
\end{equation}
where $\alpha$ and $\beta$ are hyperparameters. A high loss results in low masking (more label input), while a low loss increases masking. This adaptive mechanism allows the model to gradually transition from label-guided to vision-dominant recognition, while still being able to fall back on text input when needed.

\noindent\textbf{Final Stage: Vision-Only Prediction.}
Once the training loss falls below the threshold $\beta$, the label input is fully removed:
\begin{equation}
\tilde{\textbf{X}} = pad_{id} \cdot \mathbf{1}.
\end{equation}
At this point, the model performs standard visual recognition. Because the supervision was gradually removed, the model remains stable and robust. If the loss increases on difficult samples, masking can automatically relax in future iterations, offering dynamic feedback during training.

\subsection{Inference Phase}

During inference, TEACH does not require any ground-truth text or auxiliary input. The model structure reverts to its original encoder-decoder architecture, relying solely on the image encoder to extract visual features. These features are passed to the decoder, which autoregressively predicts the character sequence using standard decoding tokens such as [S] (start) and [P] (padding).

This design ensures that the proposed training strategy introduces no additional computation, parameters, or latency during testing. As illustrated in Figure~\ref{Fig:teach-test}, the inference process is identical to that of the original STR model, making TEACH a lightweight and plug-and-play enhancement for various architectures.

\subsection{Model Integration}

\Ours is a training-only strategy that can be seamlessly integrated into any encoder-decoder-based STR architecture. In this work, we apply it to two representative models with different decoding structures. ViTSTR~\cite{atienza2021vision_ViTSTR} uses a fully connected layer after the visual encoder for sequence prediction, while PARSeq~\cite{bautista2022PARSeq} adopts a transformer-based decoder with permuted sequence modeling. In both cases, \Ours is injected into the training pipeline without modifying the base architecture or affecting inference.

In principle, \Ours is also applicable to RNN-based architectures such as CRNN~\cite{shi2016end_CRNN}, where the CNN encoder is followed by a recurrent decoder. However, due to the sequential nature of RNNs and their limited ability to jointly leverage image and text features, our preliminary experiments found it difficult to propagate useful gradients from both branches simultaneously. As a result, we exclude CRNN-based models from the main analysis.

Similarly, large language model (LLM)-based STR frameworks like TrOCR or Clip4STR also follow encoder-decoder paradigms and can theoretically benefit from \Ours. However, such models rely on large pretrained language models and introduce substantial complexity in both training and evaluation. Since \Ours is designed as a lightweight training enhancement for vision-centric models, we focus our integration and analysis on self-contained STR architectures without external pretraining.

\section{Experiments}
\subsection{Experiments Setup}

\begin{table*}[tbp]
\centering
\scalebox{1.}{
  \begin{tabular}{cccccccccc}
  \toprule
{Method} & {Venue} & {Train Dataset}  & IIIT5K        & SVT            & IC13          & IC15           & IC15           & SVTP          & CUTE          \\
\midrule
  ASTER\cite{shi2018aster_Aster}          & PAMI’19                & MJ+ST                                                                                                            & 93.4          & 89.5           & –             & 76.1           & –              & 78.5          & 79.5          \\
  SRN\cite{yu2020towards_SRN}            & CVPR’20                & MJ+ST                                                                                                            & 94.8          & 91.5           & –             & 82.7           & –              & 85.1          & 87.8          \\
  TextScanner\cite{wan2020textscanner}    & AAAI’20                & MJ+ST                                                                                                            & 95.7          & 92.7           & 94.9          & –              & 83.5           & 84.8          & 91.6          \\
  SE-ASTER\cite{qiao2020seed_se_Aster}       & CVPR’20                & MJ+ST                                                                                                            & 93.8          & 89.6           & 92.8          & 80             & –              & 81.4          & 83.6          \\
  TRBA\cite{baek2021if_TRBA_discuss}           & CVPR’21                & MJ+ST                                                                                                            & 92.1          & 88.9           & –             & 86             & –              & 89.3          & 89.2          \\
  VisionLAN\cite{wang2021two_VisionLAN}      & ICCV’21                & MJ+ST                                                                                                            & 95.8          & 91.7           & –             & 83.7           & –              & 86            & 88.5          \\
  ABINet\cite{fang2021read_ABInet}         & CVPR’21                & MJ+ST                                                                                                            & 96.2          & 93.5           & –             & 86             & –              & 89.3          & 89.2          \\
  ViTSTR-B\cite{atienza2021vision_ViTSTR}       & ICDAR’21               & MJ+ST                                                                                                            & 88.4          & 87.7           & 92.4          & 78.5           & 72.6           & 81.8          & 81.3          \\
  ViTSTR-S\cite{atienza2021vision_ViTSTR}       & ICDAR’21               & MJ+ST                                                                                                   & 94.2         & 93.8          & 94.2        & 82.8         & 79.1           & 85.1         & 87.5          \\
  LevOCR\cite{da2022levenshtein_LevOCR}        & ECCV’22                & MJ+ST                                                                                                            & 96.6          & 92.9           & –             & 86.4           & –              & 88.1          & 91.7          \\
  MATRN\cite{na2022multi_matrn}          & ECCV’22                & MJ+ST                                                                                                            & 96.6          & 95             & 95.8          & 86.6           & 82.8           & 90.6          & \textbf{93.5} \\
  PETR\cite{wang2022petr_Petr}           & TIP’22                 & MJ+ST                                                                                                            & 95.8          & 92.4           & 97            & 83.3           & –              & 86.2          & 89.9          \\
  DiG-ViT-B\cite{yang2022reading_DiGVITB}      & MM’22                  & MJ+ST                                                                                                            & 96.7          & 94.6           & 96.9          & \textbf{87.1}           & –              & \textbf{91.0}            & 91.3          \\
  PARSeqA\cite{bautista2022PARSeq}        & ECCV’22                & MJ+ST                                                                                                            & 97            & 93.6           & 96.2          & 86.5           & 82.9           & 88.9          & 92.2          \\
 \midrule
  \rowcolor{lightgray!30}
  \textbf{ViTSTR+\Ours}       &     -               & MJ+ST                      & 94.8          & 91.3          & 95.5        & 84.8         & 80.7          & 86.8        & 89.9        \\
    \rowcolor{lightgray!30}
    \textbf{PARSeq+\Ours}           & -                   & MJ+ST                                                                                                   & \textbf{97.5} & \textbf{94.1} & \textbf{97.3}        & \textbf{87.1} & \textbf{83.4} & 90.7 & \textbf{93.8  }       \\ \bottomrule
  \end{tabular}}
\caption{
Word accuracy (\%) on six standard STR benchmarks using synthetic training data (MJ + ST). Models are evaluated on IIIT5K, SVT, IC13, IC15, SVTP, and CUTE. Our method (\Ours) improves both ViTSTR and PARSeq baselines across all benchmarks. Best results per column are in \textbf{bold}. \Ours consistently improves baseline models under synthetic training and achieves state-of-the-art performance on five out of six benchmarks.
}
 \label{tab:benchmark_synth}

\end{table*}

\begin{table*}[tbp]
\centering
\scalebox{1.}{
  \begin{tabular}{cccccccccc}
  \toprule
{Method} & {Venue} & {Train Dataset} & IIIT5K         & SVT           & IC13          & IC15           & IC15          & SVTP           & CUTE          \\ \midrule
  PARSeq+CLIPTER\cite{aberdam2023clipter_PARSeq+clipter} & ICCV’23                & N/A                                                                                         & –              & 96.6          & –             & –              & 85.9          & –              & –             \\
  DiG-ViT-B\cite{yang2022reading_DiGVITB}      & MM’22                  & Real(2.8M)                                                                                  &                & 96.5          & 97.6          & 88.9           & \textbf{–}    & 92.9           & 96.5          \\
  ViTSTR-S\cite{atienza2021vision_ViTSTR}      & ICDAR’21               & Real(3.3M)                                                                                  & 97.9           & 96            & 97.8          & 89             & 87.5          & 91.5           & 96.2          \\
  ViTSTR-S\cite{atienza2021vision_ViTSTR}      & ICDAR’21               & Real(3.3M)                                                                                            & 98.0          & 95.0         & 97.2         & 88.3          & 87.4         & 91.8          & 97.6         \\
  ABINet\cite{fang2021read_ABInet}       & CVPR’21                & Real(3.3M)                                                                                  & 98.6           & \textbf{98.2} & 98            & 90.5           & 88.7          & 94.1           & 97.2          \\
  PARSeqA\cite{bautista2022PARSeq}        & ECCV’22                & Real(3.3M)                                                                                 & 99.1           & 97.9          & \textbf{98.4} & 90.7           & \textbf{89.6}          & 95.7           & 98.3          \\
  MAERec-B\cite{jiang2023revisiting_MAERec}       & ICCV’23                & Union14M-L\cite{jiang2023revisiting_MAERec}                                                                        & 98.5           & 97.8          & 98.1          & -              & 89.5          & 94.4           & \textbf{98.6} \\ \midrule
    \rowcolor{lightgray!30}
  \textbf{ViTSTR+\Ours}       &          -              & Real(3.3M)                                                                        & 97.8           & 96.0         & 97.6         & 88.2          & 86.7         & 92.6          & 96.2         \\
    \rowcolor{lightgray!30}
  \textbf{PARSeq+\Ours}           &       -                 & Real(3.3M)                                                                                               & \textbf{99.2} & 97.5         & 98.1         & \textbf{90.8} & \textbf{89.6} & \textbf{96.0} & 97.6         \\ \bottomrule
  \end{tabular}}
\caption{
Word accuracy (\%) on six STR benchmarks using real-world training data (3.3M images). PARSeq+\Ours achieves new state-of-the-art results on five out of six benchmarks, outperforming strong pretrained models including DiG-ViT, ABINet, and MAERec. Best results per column are in \textbf{bold}. \Ours leads to strong performance gains even in high-resource settings, surpassing prior SOTA methods without introducing inference overhead.
}
 \label{tab:benchmark_real}
\end{table*}

\begin{table*}[t]
\centering
\scalebox{1.}{
  \begin{tabular}{cccccc}
  \toprule
{Method} & {Train data} & ArT       & COCO      & Uber     & Total \\  \midrule
  CRNN                    & MJ+ST                       & 57.3±0.1  & 49.3±0.6  & 33.1±0.3 & 41.1±0.3 \\
  ViTSTR-S                & MJ+ST                       & 66.1±0.1  & 56.4±0.5  & 37.6±0.3 & 47.0±0.2 \\
  TRBA                    & MJ+ST                       & 68.2±0.1  & 61.4±0.4  & 38.0±0.3 & 48.3±0.2 \\
  ABINet                  & MJ+ST                       & 65.4±0.4  & 57.1±0.8  & 34.9±0.3 & 45.2±0.3 \\
  PARSeqN                 & MJ+ST                       & 69.1±0.2  & 60.2±0.8  & 39.9±0.5 & 49.7±0.3 \\
  PARSeqA                 & MJ+ST                       & 70.7±0.1  & 64.0±0.9  & 42.0±0.5 & 51.8±0.4 \\ \midrule
    \rowcolor{lightgray!30}
      \textbf{ViTSTR+\Ours}           & MJ+ST                       & 66.7±0.2  & 56.2±1.3  & 38.4±0.3 & 47.7±0.1 \\
          \rowcolor{lightgray!30}
  \textbf{PARSeq+\Ours}           & MJ+ST                       & \textbf{71.0±0.1}  & \textbf{64.6±0.1}  & \textbf{43.0±0.3} & \textbf{52.5±0.1} \\
 \midrule
  CRNN                    & Real(3.3M)                  & 66.8±0.2  & 62.2±0.3  & 51.0±0.2 & 56.3±0.2 \\
  ViTSTR-S                & Real(3.3M)                  & 81.1±0.1  & 74.1±0.4  & 78.2±0.1 & 78.7±0.1 \\
  TRBA                    & Real(3.3M)                  & 82.5±0.2  & 77.5±0.2  & 81.2±0.3 & 81.3±0.2 \\
  ABINet                  & Real(3.3M)                  & 81.2±0.1  & 76.4±0.1  & 71.5±0.7 & 74.6±0.4 \\
  PARSeqN                 & Real(3.3M)                  & 83.0±0.2  & 77.0±0.2  & 82.4±0.3 & 82.1±0.2 \\
  PARSeqA                 & Real(3.3M)                  & 84.5±0.1  & 79.8±0.1  & 84.5±0.1 & 84.1±0.0 \\ \midrule
    \rowcolor{lightgray!30}
  \textbf{ViTSTR+\Ours}           & Real(3.3M)                  & 81.9±0.1 & 75.3±0.4 & 79.8±0.2   & 79.5±0.1 \\
    \rowcolor{lightgray!30}
      \textbf{PARSeq+\Ours}           & Real(3.3M)                  & \textbf{85.0±0.2} & \textbf{80.7±0.2} & \textbf{85.6±0.5 }  & \textbf{84.2±0.4} \\
\bottomrule
  
  \end{tabular}}
\caption{
Word accuracy (\%) on six STR benchmarks using real-world training data (3.3M images). PARSeq+\Ours achieves new state-of-the-art results on five out of six benchmarks, outperforming strong pretrained models including DiG-ViT, ABINet, and MAERec. Best results per column are in \textbf{bold}.
\Ours leads to strong performance gains even in high-resource settings, surpassing prior SOTA methods without introducing inference overhead.
}
  \label{tab:hard_sets}
\end{table*}

Following standard practice in scene text recognition (STR)~\cite{baek2021if_TRBA_discuss, bautista2022PARSeq}, we adopt both synthetic and real-world datasets for training and evaluation.

\noindent\textbf{Synthetic datasets.}  
We use two widely adopted synthetic datasets to pretrain STR models: MJSynth (MJ)~\cite{jaderberg2014synthetic_MJ}, containing approximately 9 million rendered word images, and SynthText (ST)~\cite{gupta2016synthetic_ST}, which provides around 6.9 million synthetically generated text samples embedded in natural scenes. These datasets serve as the standard foundation for training baseline models and for comparison with prior work.

\noindent\textbf{Real-world datasets.}  
To further validate performance in practical settings, we use multiple large-scale real datasets: COCO-Text~\cite{veit2016coco_Coco}, RCTW17~\cite{shi2017icdar2017_RCTW17}, Uber-Text~\cite{zhang2017uber_Uber}, ArT~\cite{chng2019icdar2019_art}, LSVT~\cite{sun2019icdar_LSVT}, MLT19~\cite{nayef2019icdar2019_MLT19}, and ReCTS~\cite{zhang2019icdar_ReCTS}. A detailed taxonomy of these datasets is available in~\cite{baek2021if_TRBA_discuss}.  
In addition, we include two recent datasets built on OpenImages~\cite{krasin2017openimages_openimages}: TextOCR~\cite{singh2021textocr_TextOCR} and annotations from the OpenVINO toolkit~\cite{krylov2021open_openvino}, both of which represent diverse and challenging real-world distributions.

\noindent\textbf{Evaluation benchmarks.}  
For evaluation, we follow prior work and adopt six widely used benchmark datasets: IIIT 5k-word (IIIT5k)~\cite{mishra2012scene_iii5k}, Street View Text (SVT)~\cite{wang2011end_SVt}, SVT-Perspective (SVTP)~\cite{phan2013recognizing_svtp}, ICDAR 2013 (IC13)~\cite{karatzas2013icdar_IC13}, ICDAR 2015 (IC15)~\cite{karatzas2015icdar_IC15}, and CUTE80 (CUTE)~\cite{risnumawan2014robust_cute80}. Following~\cite{long2020unrealtext_sensitive4test}, we use case-sensitive annotations for IIIT5k, SVT, SVTP, and CUTE.  

Notably, IC13 and IC15 are reported in two common variants—IC13 includes either 857 or 1,015 test samples, while IC15 includes either 1,811 or 2,077 samples. To standardize terminology, we refer to the combination of IIIT5k, SVT, SVTP, CUTE, IC13 (1,015), and IC15 (2,077) as the \textbf{ALL} benchmark set, totaling 7,672 images. In contrast, we refer to the configuration using IC13 (857) and IC15 (1,811) as the \textbf{Clean} subset, consistent with prior work~\cite{bautista2022PARSeq}.

\subsection{Implementation Details}

\noindent\textbf{\Ours variants.}
\Ours is integrated after the image encoding stage of each STR model. The injected label tensor $\textbf{X}_{\text{input}} \in \mathbb{R}^{B \times S \times E_s}$ is dynamically masked based on model loss, as defined in \cref{eq:mask_ratio}, where $B$ is the batch size, $S$ is the fixed maximum sequence length (set to 25 unless otherwise specified), and $E_s$ is the embedding dimension. To determine masking parameters $\alpha$ and $\beta$, we first pretrain the baseline model without \Ours until convergence. We then set $\beta$ slightly below the stabilized loss value and select $\alpha$ to control the masking decay rate: smaller $\alpha$ values accelerate masking when convergence is fast, while larger values slow the decay for more gradual guidance. When $r \leq 0$, the label input is fully masked.

\noindent\textbf{Traning details.}
For all models, we follow the official training configurations from PARSeq~\cite{bautista2022PARSeq}, ViTSTR~\cite{atienza2021vision_ViTSTR}, and CRNN~\cite{shi2016end_CRNN}, unless otherwise noted. Due to the introduction of label embeddings, models with \Ours require slightly more training iterations to converge. To ensure fair comparison, all models (with and without \Ours) are trained for the same number of epochs, determined by the stabilization point of the most stable configuration. Experiments are conducted on a dual NVIDIA V100 GPU. Our code and training configurations will be released upon publication.


\begin{table}[t]
    \centering
    \scalebox{0.8}{
    \begin{tabular}{c|c|c|c}
        \toprule
        \textbf{Original Image} & \textbf{Model} & \textbf{Prediction} & \textbf{Ground Truth} \\
        \midrule
        \multirow{4}{*}{\includegraphics[width=2.5cm]{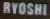}} 
        & \textbf{PARSeq-\Ours} & \texttt{\textbf{ryoshi}} & \multirow{4}{*}{\texttt{ryoshi}} \\
        & PARSeq & \texttt{r\textcolor{red}{v}oshi} & \\
        & \textbf{ViTSTR-\Ours} & \texttt{\textbf{ryoshi}} & \\
        & ViTSTR & \texttt{\textcolor{red}{a}yoshi} & \\
        \midrule
        
        \multirow{4}{*}{\includegraphics[width=2.5cm]{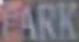}} 
        & \textbf{PARSeq-\Ours} & \texttt{\textbf{park}} & \multirow{4}{*}{\texttt{park}} \\
        & PARSeq & \texttt{\textcolor{red}{l}ark} & \\
        & \textbf{ViTSTR-\Ours} & \texttt{\textbf{park}} & \\
        & ViTSTR & \texttt{\textcolor{red}{e}ark} & \\
        \midrule
        
        \multirow{4}{*}{\includegraphics[width=2.5cm]{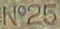}} 
        & \textbf{PARSeq-\Ours} & \texttt{\textbf{no25}} & \multirow{4}{*}{\texttt{no25}} \\
        & PARSeq & \texttt{no\textcolor{red}{5}5} & \\
        & \textbf{ViTSTR-\Ours} & \texttt{\textbf{no25}} & \\
        & ViTSTR & \texttt{n\textcolor{red}{$\_$}25} & \\
        \midrule
        
        \multirow{4}{*}{\includegraphics[width=2.5cm]{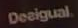}} 
        & \textbf{PARSeq-\Ours} & \texttt{\textbf{desigual}} & \multirow{4}{*}{\texttt{desigual}} \\
        & PARSeq & \texttt{d\textcolor{red}{ee}igual} & \\
        & \textbf{ViTSTR-\Ours} & \texttt{\textbf{desigual}} & \\
        & ViTSTR & \texttt{d\textcolor{red}{ea}igual} & \\
        \midrule

        \multirow{4}{*}{\includegraphics[width=2.5cm]{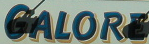}} 
        & \textbf{PARSeq-\Ours} & \texttt{\textbf{galore}} & \multirow{4}{*}{\texttt{galore}} \\
        & PARSeq & \texttt{\textcolor{red}{c}alore} & \\
        & \textbf{ViTSTR-\Ours} & \texttt{\textbf{galore}} & \\
        & ViTSTR & \texttt{\textcolor{red}{c}alore} & \\
        \midrule
        
        \multirow{4}{*}{\includegraphics[width=2.5cm]{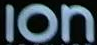}} 
        & \textbf{PARSeq-\Ours} & \texttt{\textbf{ion}} & \multirow{4}{*}{\texttt{ion}} \\
        & PARSeq & \texttt{\textcolor{red}{l}on} & \\
        & \textbf{ViTSTR-\Ours} & \texttt{\textbf{ion}} & \\
        & ViTSTR & \texttt{\textcolor{red}{l}on} & \\
        \midrule
        
        \multirow{4}{*}{\includegraphics[width=2.5cm]{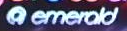}} 
        & \textbf{PARSeq-\Ours} & \texttt{\textbf{emerold}} & \multirow{4}{*}{\texttt{emerold}} \\
        & PARSeq & \texttt{emer\textcolor{red}{a}ld} & \\
        & \textbf{ViTSTR-\Ours} & \texttt{emer\textcolor{red}{ra}d} & \\
        & ViTSTR & \texttt{em\textcolor{red}{eeaa}ld} & \\
        \midrule
        
        \multirow{4}{*}{\includegraphics[width=2.5cm]{images/sample_images/image-000002844.png}} 
        & \textbf{PARSeq-\Ours} & \texttt{\textbf{centre}} & \multirow{4}{*}{\texttt{centre}} \\
        & PARSeq & \texttt{cent\textcolor{red}{ie}} & \\
        & \textbf{ViTSTR-\Ours} & \texttt{\textbf{centre}} & \\
        & ViTSTR & \texttt{cent\textcolor{red}{ie}} & \\
        
        \bottomrule
    \end{tabular}}
  \caption{
Qualitative comparison of recognition results across different models. 
Each row shows the original input image, predictions from baseline models (PARSeq and ViTSTR), and our \Ours-enhanced versions. 
\Ours improves accuracy on challenging cases, such as distorted text, curved styles, and low-contrast inputs.
}
\label{tab:qualitative_comparison}
\end{table}

\subsection{Results}

\noindent\textbf{Effect of \Ours.}
We first evaluate the impact of \Ours by comparing its performance on two representative STR models: ViTSTR and PARSeq. As shown in Table~\ref{tab:benchmark_synth}, \Ours brings consistent improvements across all six standard benchmarks when trained on synthetic datasets (MJ + ST). Specifically, ViTSTR+\Ours improves upon the ViTSTR baseline by up to 1.7\% on IC13 and 1.6\% on CUTE. PARSeq+\Ours shows even more substantial gains, achieving 97.5\% on IIIT5K and 93.4\% on CUTE, outperforming the original PARSeq by 0.5--1.2\% across most benchmarks.
Similar trends are observed when training with real data (3.3M real images). As reported in Table~\ref{tab:benchmark_real}, ViTSTR+\Ours improves on IC15 and SVTP by 0.6\% and 0.8\%, respectively, while PARSeq+\Ours pushes performance beyond 99.0\% on IIIT5K and above 97.6\% on CUTE, showing the effectiveness of our method in high-resource settings. These consistent improvements validate that \Ours serves as a universal training enhancement and is effective across model architectures and data domains.

\noindent\textbf{Comparison on Synthetic Benchmarks.}
Table~\ref{tab:benchmark_synth} presents word accuracy on six widely-used benchmarks under synthetic training (MJ + ST). Our method, PARSeq+\Ours, achieves the highest accuracy on five out of six datasets, including IIIT5k (97.5\%), IC13 (97.3\%), IC15 (83.4\%), SVTP (90.7\%), and CUTE (93.4\%). TEACH enhances both PARSeq and ViTSTR baselines, reflecting its effectiveness in guiding early training. Notably, TEACH outperforms strong transformer-based methods like DiG-ViT~\cite{yang2022reading_DiGVITB}, PETR~\cite{wang2022petr_Petr}, and MATRN~\cite{na2022multi_matrn}, despite not using external pretraining or language models.

\noindent\textbf{Comparison on Real-World Benchmarks.}
As shown in Table~\ref{tab:benchmark_real}, under real-data training (3.3M samples), PARSeq+\Ours again achieves top performance across most benchmarks. It attains state-of-the-art results on IC15 (90.8\%), SVTP (96.0\%), and CUTE (97.6\%), surpassing recent methods like MAERec~\cite{jiang2023revisiting_MAERec} and ABINet~\cite{fang2021read_ABInet}. ViTSTR+\Ours also demonstrates consistent improvements, indicating that TEACH generalizes across architectures. These results confirm that TEACH remains highly effective under real-world training conditions.

\noindent\textbf{Comparison on Challenging Large-Scale Benchmarks.}
Table~\ref{tab:hard_sets} reports accuracy on ArT, COCO-Text, and UberText—datasets characterized by complex layout, noise, and distortions. TEACH-enhanced models show strong robustness, with PARSeq+\Ours outperforming all other models and achieving the best overall accuracy (84.2\%). Particularly on UberText, which includes diverse and low-quality samples, PARSeq+\Ours yields a 2.0\% gain over its baseline. These results highlight TEACH's capability to handle challenging, real-world scenarios where standard STR systems often fail.

\noindent\textbf{Qualitative Results Comparison.}
To better understand how \Ours improves scene text recognition, we compare qualitative results from baseline models (PARSeq and ViTSTR) and their \Ours-enhanced variants. As shown in Table~\ref{tab:qualitative_comparison}, \Ours consistently improves prediction quality, especially on challenging samples that contain blur, perspective distortion, non-standard fonts, or low contrast.
For instance, in the first and fourth rows, both baselines misrecognize characters due to unusual fonts or visual clutter, while \Ours correctly predicts the full word. In the case of "emerald" and "centre," \Ours significantly reduces errors caused by partial occlusions or ambiguous shapes. These results demonstrate that progressive label injection during training helps the model form stronger visual representations and reduces its sensitivity to noise.
The examples confirm that \Ours not only improves average performance but also enhances robustness in real-world complex scenes—an essential quality for practical STR applications.

\begin{figure}[htbp]
  \centering
  \begin{subfigure}[b]{0.45\textwidth}
    \centering
    \includegraphics[scale=0.23]{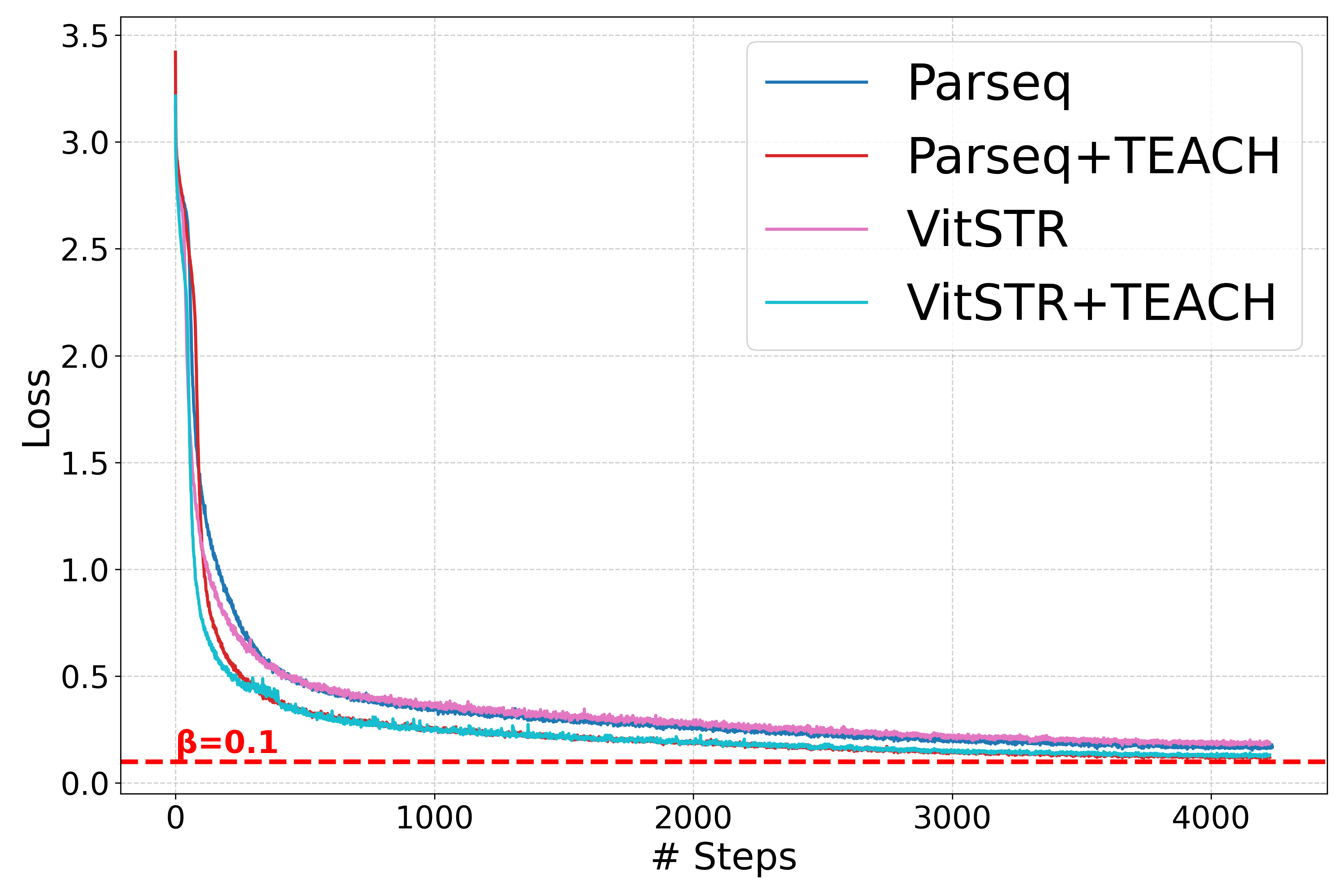}
    \caption{Overall Convergence}
    \label{subfig:overall}
  \end{subfigure}
  \\
  \begin{subfigure}[b]{0.45\textwidth}
    \centering
    \includegraphics[scale=0.23]{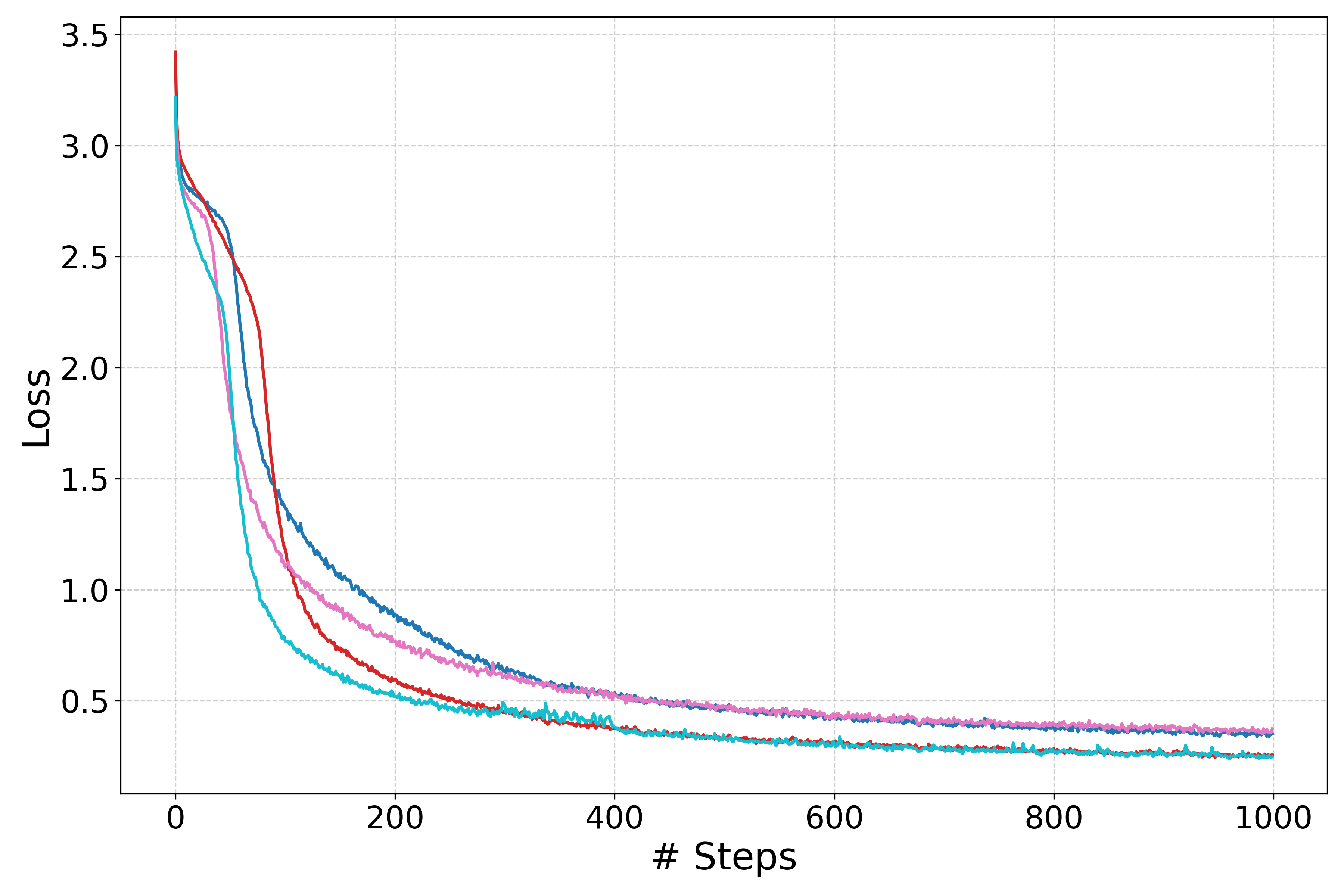}
    \caption{Early-stage Convergence}
    \label{subfig:first1000}
  \end{subfigure}
  \hfill
  \begin{subfigure}[b]{0.45\textwidth}
    \centering
    \includegraphics[scale=0.23]{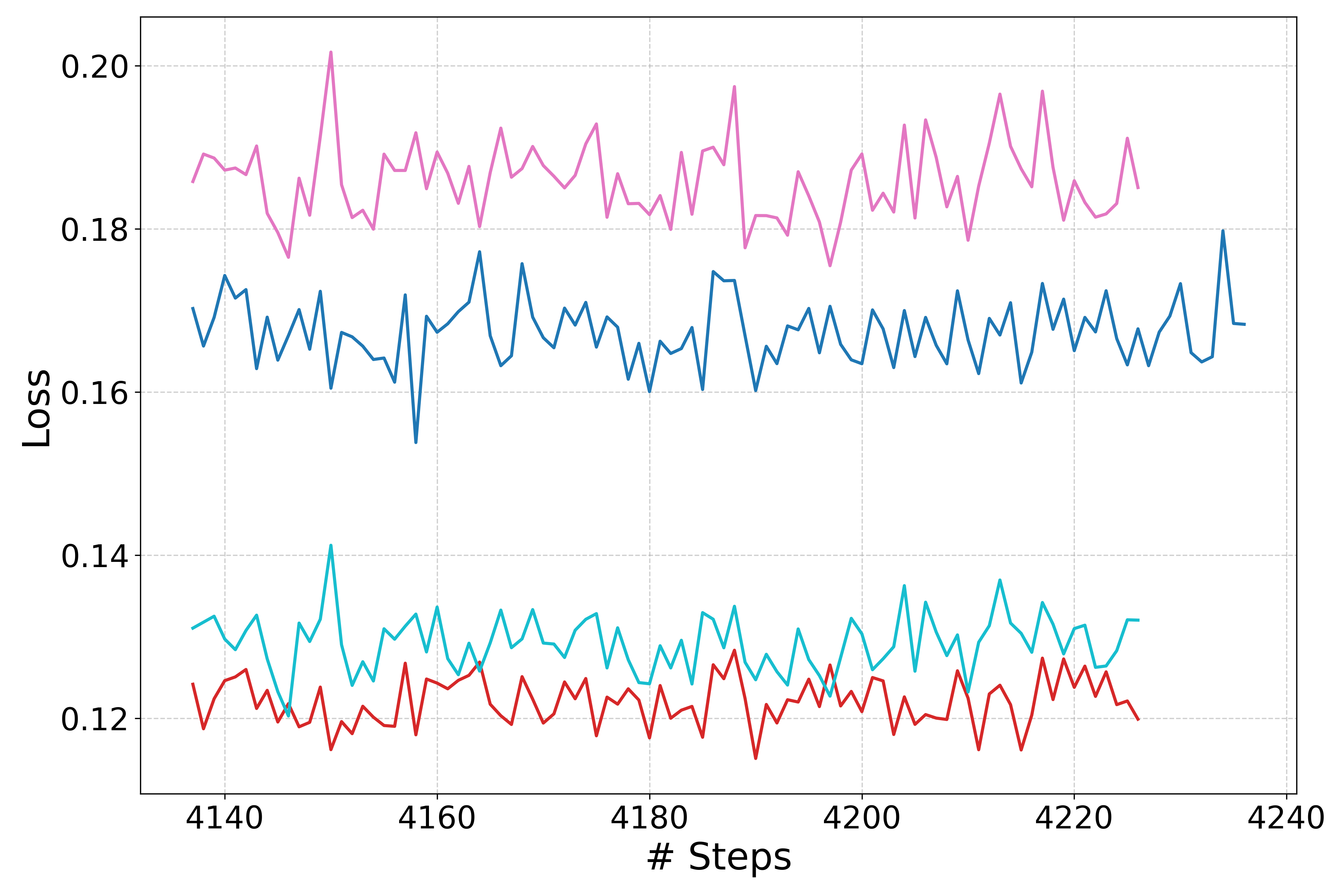}
    \caption{Final-stage Stability}
    \label{subfig:last100}
  \end{subfigure}
\caption{
Comparison of loss function convergence on the training set for different models. 
(a) Full training loss curve across 50 epochs. 
(b) Early-stage loss within the first 1000 iterations, showing accelerated convergence with \Ours. 
(c) Final-stage loss in the last 100 iterations, where \Ours achieves lower terminal loss. 
All plots include inline annotations for clarity. 
In this experiment, $\alpha = 2$ and $\beta = 0.1$.
}
  \label{Fig:loss_convergence}
\end{figure}




\subsection{Ablation Analysis}

To better understand the contribution of each component in \Ours, we conduct ablation studies on the IIIT5K, IC13, and CUTE benchmarks under the MJ + ST training setting. We focus on the effect of progressive masking, as well as the impact of label injection alone.

\begin{table}[t]
\centering
\scalebox{0.8}{
\begin{tabular}{lcccc}
\toprule
\multirow{2}{*}{\textbf{Method}} & \multicolumn{2}{c}{\textbf{Real}} & \multicolumn{2}{c}{\textbf{Synthetic}} \\
\cmidrule(lr){2-3} \cmidrule(lr){4-5}
& PARSeq & ViTSTR & PARSeq & ViTSTR \\
\midrule
Baseline                     & 96.00 & 94.31 & 92.92 & 88.60 \\
\ + Linear Masking  & 96.39 & 94.59 & 93.16 & 90.14 \\
\ + Loss-based Masking (ours)        & \textbf{96.43} & \textbf{95.36} & \textbf{93.60} & \textbf{90.14} \\
\bottomrule
\end{tabular}}
\caption{
Comparison of different masking schedule strategies in \Ours using PARSeq and ViTSTR under synthetic (MJ+ST) and real (3.3M) data. The loss-based masking consistently outperforms the fixed linear decay, showing better adaptability to model convergence.
}
\label{tab:masking_schedule}
\end{table}

\noindent\textbf{Effect of Masking Schedules.}
We further compare different strategies for controlling the masking ratio in \Ours. Specifically, we test (1) a linear decay schedule where the ratio is linearly reduced based on the global step, and (2) the proposed loss-aware dynamic masking scheme that adjusts the ratio according to recent loss values.
Results are shown in Table~\ref{tab:masking_schedule}. Across both PARSeq and ViTSTR backbones, and under both real and synthetic training regimes, the loss-based schedule consistently achieves higher accuracy than both the baseline and the linear decay variant. While linear masking shows modest improvements over the baseline, its fixed nature may result in suboptimal supervision timing. In contrast, the loss-aware approach adapts to model convergence speed and maintains label guidance when needed, leading to more stable and effective learning. This confirms that dynamic supervision is key to the success of \Ours.

\noindent\textbf{Impact of $\alpha$ and $\beta$ Hyperparameters.}
The dynamic masking ratio in TEACH is controlled by the loss-adaptive formula defined in Equation~\ref{eq:mask_ratio}, where $\alpha$ determines sensitivity to loss fluctuations and $\beta$ sets the convergence threshold. To investigate the robustness of TEACH under different schedules, we conduct an ablation study on combinations of $\alpha \in \{0.5, 1, 2\}$ and $\beta \in \{0.01, 0.05, 0.1\}$ using the PARSeq backbone.
Table~\ref{tab:ablation_alpha_beta} summarizes results on both synthetic and real training data. We observe that TEACH consistently improves over the baseline across most settings. In particular, a larger $\alpha$ combined with a higher $\beta$ (i.e., slower masking decay) tends to yield better results, with the best performance achieved when $\alpha/\beta = 2/0.1$. This configuration enables prolonged access to label supervision during early training while still ensuring full masking in later stages.

\begin{table}[h]
\centering
\begin{tabular}{ccccc}
\toprule
\textbf{Data Type} & \textbf{$\alpha$} & \textbf{$\beta=0.1$} & \textbf{$\beta=0.05$} & \textbf{$\beta=0.01$} \\
\midrule
\multirow{3}{*}{Real} 
  & 2   & \textbf{96.43} & 96.29 & \textbf{96.43} \\
  & 1   & 96.33 & \textbf{96.33} & 96.27 \\
  & 0.5 & 96.39 & 96.32 & 93.08 \\
\midrule
\multirow{3}{*}{Synth} 
  & 2   &\textbf{93.24} & \textbf{93.60} & \textbf{93.50} \\
  & 1   & 92.92 & 93.42 & 93.24 \\
  & 0.5 & 93.12 & 92.87 & 93.16 \\
\bottomrule
\end{tabular}
\caption{Ablation study of TEACH on different combinations of $\alpha$ (rows) and $\beta$ (columns) for both real and synthetic datasets using the PARSeq model. Accuracy is averaged across six STR benchmarks.}
\label{tab:ablation_alpha_beta}
\end{table}

\begin{table}[t]
\centering
\begin{tabular}{lcc}
\toprule
\textbf{Model} & \textbf{Real} & \textbf{Synthetic} \\
\midrule
CRNN Baseline   & 91.19 & 86.13 \\
CRNN + TEACH    & \textbf{91.27} & \textbf{86.56} \\
\bottomrule
\end{tabular}
\caption{Effect of integrating TEACH into the CRNN architecture. Accuracy is evaluated on both synthetic and real datasets.}
\label{tab:crnn_xinput}
\end{table}

\textbf{CRNN Integration Analysis.}
To verify the general applicability of TEACH beyond transformer-based models, we integrate it into the convolutional-recurrent CRNN architecture. In this setup, label embeddings are concatenated with the CNN output before being passed to the RNN decoder. As shown in Table~\ref{tab:crnn_xinput}, this integration yields consistent performance improvements across both real and synthetic training settings, despite the inherent sequential bottleneck of RNN decoders. This result supports the generality of TEACH as a training-time optimization method.

\noindent\textbf{Training Dynamics and Convergence Behavior.}
To investigate how \Ours affects the optimization dynamics, we compare the training loss curves of different models with and without our method. As shown in \cref{Fig:loss_convergence}, models trained with \Ours exhibit faster convergence in the early stages (\cref{subfig:first1000}) and achieve lower final loss values in the late stages (\cref{subfig:last100}). This demonstrates that the progressive label injection strategy provides effective guidance during early training, while the gradual masking enables a smooth transition to vision-only prediction.
Assuming similar data distributions between training and testing, lower training loss typically correlates with higher accuracy. Thus, the improved convergence profile further confirms the effectiveness of \Ours as a training-time optimization strategy.

\section{Conclusion}

In this work, we proposed \Ours, a lightweight and model-agnostic training strategy for scene text recognition (STR). \Ours injects ground-truth labels into the input sequence during training and gradually masks them based on model performance. This progressive supervision simulates a curriculum learning process, enabling the model to shift from explicit label guidance to fully vision-based recognition without relying on language models or additional inference cost.
We demonstrated that \Ours can be seamlessly integrated into existing encoder-decoder architectures, including both transformer-based and fully connected decoders. Extensive experiments on synthetic and real-world datasets show that \Ours consistently improves recognition accuracy and enhances model robustness, especially under challenging conditions. Compared with state-of-the-art STR methods, our approach achieves superior performance across standard benchmarks and large-scale evaluation sets.
\Ours offers a general framework for training-time guidance and opens up new directions for lightweight supervision in vision tasks. Future work includes exploring its application in multilingual STR, handwritten text recognition, and extending the progressive supervision paradigm to other sequence modeling tasks.


\end{document}